\documentclass[manuscript]{acmart}

\newif\ifapx

\newif\ifgenplot

\newcommand{\ourmaintitle}{Causal Inference by Stochastic Complexity}

\newcommand{\ourmethod}{\textsc{cisc}\xspace}
\newcommand{\oururl}{\url{http://eda.mmci.uni-saarland.de/cisc/}}

\usepackage{latexsym}
\usepackage[ruled, vlined, nofillcomment, linesnumbered]{algorithm2e}
\usepackage{eda} 
\usepackage{bm}	
\usepackage{enumitem} 
\usepackage{booktabs} 
\usepackage[labelfont=sf, textfont=sf,singlelinecheck=off,justification=centering]{subcaption} 
\usepackage{footmisc} 
\usepackage{balance} 

\usepgfplotslibrary{fillbetween}

\graphicspath{ {./} }

\ifpdf
\usetikzlibrary{external}
\fi

\newcommand{\xtoy}{\ensuremath{X \rightarrow Y}\xspace}
\newcommand{\ytox}{\ensuremath{Y \rightarrow X}\xspace}
\newcommand{\cause}{\ensuremath{\mathit{cause}}\xspace}
\newcommand{\effect}{\ensuremath{\mathit{effect}}\xspace}
\newcommand{\mclass}{\ensuremath{\mathcal{M}}\xspace}
\newcommand{\scomp}[1]{\ensuremath{\mathcal{S}(#1)}\xspace}
\newcommand{\scompxtoy}{\ensuremath{\mathcal{S}_{\xtoy}}\xspace}
\newcommand{\scompytox}{\ensuremath{\mathcal{S}_{\ytox}}\xspace}
\newcommand{\params}{\ensuremath{\bm{\theta}}\xspace}
\newcommand{\paramsspace}{\ensuremath{\Theta}\xspace}
\newcommand{\pnml}[1]{\ensuremath{P_{\textsc{NML}}(#1)}\xspace}
\newcommand{\mnmclass}{\ensuremath{\mathcal{M}_m}\xspace}
\newcommand{\spacex}{\ensuremath{\mathcal{X}}\xspace}
\newcommand{\spacey}{\ensuremath{\mathcal{Y}}\xspace}

\newcommand{\indep}{\ensuremath{\perp \! \! \! \perp}\xspace}
\newcommand{\bigo}[1]{\ensuremath{\mathcal{O}(#1)\xspace}}
\newcommand{\peq}{\ensuremath{\stackrel{+}{=}}\xspace}

\newcommand{\dc}{\textsc{dc}\xspace}
\newcommand{\dr}{\textsc{dr}\xspace}

\newtheorem{postulate}{Postulate}

\setcopyright{rightsretained}
\acmDOI{10.475/123_4}
\acmISBN{123-4567-24-567/08/06}
\acmConference[KDD'17]{ACM SIGKDD conference}{August 2017}{Halifax, Canada} 
\acmYear{2017}\copyrightyear{2017}\acmPrice{15.00}

\begin{document}
\setlength{\pdfpagewidth}{8.5in}
\setlength{\pdfpageheight}{11in}

\title{\ourmaintitle}

\author{Kailash Budhathoki}
\affiliation{%
	\institution{Max Planck Institute for Informatics and Saarland University, Germany}
}
\email{kbudhath@mpi-inf.mpg.de}

\author{Jilles Vreeken}
\affiliation{%
	\institution{Max Planck Institute for Informatics and Saarland University, Germany}
}
\email{jilles@mpi-inf.mpg.de}

\begin{abstract}

The algorithmic Markov condition states that the most likely causal 
direction between two random variables $X$ and $Y$ can be identified 
as that direction with the lowest Kolmogorov complexity. 
Due to the halting problem, however, this notion is not computable. 

We hence propose to do causal inference by stochastic complexity. 
That is, we propose to approximate Kolmogorov complexity via the 
Minimum Description Length (MDL) principle, using a score that is 
mini-max optimal with regard to the model class under consideration. 
This means that even in an adversarial setting, such as when the 
true distribution is not in this class, we still obtain the optimal 
encoding for the data relative to the class.

We instantiate this framework, which we call \ourmethod, 
for pairs of univariate discrete variables,
using the class of multinomial distributions. 
Experiments show that \ourmethod is highly accurate on 
synthetic, benchmark, as well as real-world data, 
outperforming the state of the art by a margin, 
and scales extremely well with regard to sample and domain sizes.
 
\end{abstract}

\ccsdesc[500]{Mathematics of computing~Causal networks}
\ccsdesc[500]{Mathematics of computing~Information theory}
\ccsdesc[500]{Information systems~Data mining}

\keywords{Causal Inference, Stochastic Complexity, NML}

\maketitle

\section{Introduction}\label{sec:introduction}

Causal inference from observational data---that is, identifying cause and effect in data that was not collected through carefully controlled randomised trials---is a fundamental problem in both business and science~\cite{spirtes:00:book,pearl:09:book}.
A particularly interesting setting is to tell cause from effect between a pair of random variables $X$ and $Y$, 
given data over the joint distribution. That is, to identify which of $\xtoy$ or $\ytox$ is the most likely causal direction. 

In recent years, a number of important ideas have been proposed that allow for accurate causal inference based on properties of the joint distribution. These ideas include that of the Additive Noise Model (ANM), where we assume the effect is a function of the cause with additive noise independent of the cause~\cite{shimizu:06:anm,peters:10:discreteanm,peters:14:continuousanm}, and that of the algorithmic Markov condition~\cite{janzing:10:algomarkov,budhathoki:16:origo} which is based on Kolmogorov Complexity. Loosely speaking, the key idea is that if $X$ causes $Y$, the shortest description of the joint distribution $P(X,Y)$ is given by the separate descriptions of $P(X)$ and $P(Y\mid X)$. That is, if \xtoy, these two distributions will be less dependent than $P(Y)$ and $P(X \mid Y)$. 
However, as Kolmogorov complexity is not computable, any method using this observation requires a computable approximation of this notion, which in general involves arbitrary choices~\cite{sgouristsa:15:cure,vreeken:15:ergo,liu:16:dc,janzing:12:igci}.

In this paper, for the first time, we define a causal inference rule based on the algorithmic Markov condition using stochastic complexity. More in particular, we approximate Kolmogorov complexity via the Minimum Description Length (MDL) principle using a score that is mini-max optimal with regard to the model class under consideration. This means that even if the true data generating distribution does not reside in the model class \mclass under consideration, we still obtain the optimal encoding for the data relative to \mclass~\cite{grunwald:07:book}. Best of all, unlike Kolmogorov complexity, stochastic complexity is \emph{computable}.

We show the strength of this approach by instantiating it for pairs of univariate discrete data using the class of multinomials. For this class the stochastic complexity is computable remarkably efficiently, by which our score has only a linear-time computational complexity. Through experiments we show that our method, \ourmethod, for causal inference by stochastic complexity, performs very well in practice. The strength of the mini-max property shows when we consider synthetic data where we vary the data generating process---\ourmethod outperforms the state of the art by a margin, including for out-of-model distributions such as geometric, hypergeometric, and Poisson. On the T\"{u}bingen benchmark data set of 95 univariate pairs, \ourmethod significantly outperforms the existing proposals for discrete data, with an accuracy of 100\% over the 21 pairs it is most certain about, and an overall accuracy of $67\%$, which is comparable to the state of the art for causal inference on continuous-valued data. Last, but not least, we perform three case studies which show \ourmethod indeed infers sensible causal directions from real-world data.

In sum, the main contributions of this paper are as follows.
\begin{itemize}[noitemsep,topsep=2pt]
	\item[(a)] we propose the first computable framework for causal inference by the algorithmic Markov condition with provable mini-max optimality guarantees,
	\item[(b)] define a causal indicator for pairs of discrete variables based on stochastic complexity,
	\item[(c)] show how to efficiently compute it,
	\item[(d)] provide extensive experimental results on synthetic, benchmark, and real-world data, and
	\item[(e)] make our implementation and all used data available
\end{itemize}

The paper is structured as usual. We introduce notation and give preliminaries in Sec.~\ref{sec:prelim}, and give a brief primer to causal inference by Kolmogorov complexity in Sec.~\ref{sec:foundation}. We present \ourmethod, our practical instantiation based on stochastic complexity score in Sec.~\ref{sec:practical}. Related work is discussed in Sec.~\ref{sec:rel}, and we evaluate \ourmethod empirically in Sec.~\ref{sec:exps}. We round up with discussion in Sec.~\ref{sec:disc} and conclude in Sec.~\ref{sec:concl}.

\section{Preliminaries}\label{sec:prelim}
In this section, we introduce notations and background definitions we will use in subsequent sections.
\subsection{Kolmogorov Complexity}
The Kolmogorov complexity of a finite binary string $x$ is the length of the shortest binary program $p^*$ for a Universal Turing machine $\mathcal{U}$ that generates $x$, and then halts~\cite{kolmogorov:65:information, vitanyi:93:book}. Formally, we have
\[
K(x) = \min \left \{  |p| : p \in \{0,1\}^*, \mathcal{U}(p) = x \right \} \; .
\]
Simply put, $p^*$ is the most succinct \emph{algorithmic} description of $x$, and the Kolmogorov complexity of $x$ is the length of its ultimate lossless compression. Conditional Kolmogorov complexity, $K(x \mid y) \leq K(x)$, is then the length of the shortest binary program $p^*$ that generates $x$, and halts, given $y$ as input. 

The amount of \emph{algorithmic information} contained in $y$ about $x$ is $I(y:x) = K(y) - K(y \mid x^*)$, where $x^*$ is the shortest binary program for $x$, defining $I(x:y)$ analogously. Intuitively, it is the number of bits that can be saved in the description of $y$ when the shortest description of $x$ is already known. Algorithmic information is symmetric, i.e.\ $I(y:x) \peq I(x:y)$, where \peq denotes equality up to an additive constant, and therefore also called \emph{algorithmic mutual information}~\cite{vitanyi:93:book}. Two strings $x$ and $y$ are \emph{algorithmically independent} if they have no algorithmic mutual information, i.e.\ $I(x:y) \peq 0$.

For our purpose, we also need the Kolmogorov complexity of a distribution.  The Kolmogorov complexity of a probability distribution $P$, $K(P)$, is the length of the shortest program that outputs $P(x)$ to precision $q$ on input $\langle x, q \rangle$~\cite{grunwald:08:ait}. More formally, we have
\[
K(P) = \min \left \{ |p| : p \in \{0,1\}^*, |\mathcal{U}(\langle x, \langle q, p \rangle \rangle) - P(x) | \leq 1/q \right \} \; .
\]

We refer the interested reader to Li \& Vit{\'a}nyi~\cite{vitanyi:93:book} for more
details on Kolmogorov complexity.

\section{Causal Inference by Complexity}\label{sec:foundation}
Given two correlated variables $X$ and $Y$, we are interested in inferring their causal relationship. In particular, we want to infer whether $X$ causes $Y$, whether $Y$ causes $X$, or they are only correlated. In doing so, we assume causal sufficiency. That is, there is no confounding variable, i.e.\ hidden common cause $Z$ of $X$ and $Y$. We use $\xtoy$ to indicate $X$ causes $Y$.

We base our causal inference method on the following postulate:

\begin{postulate}[independence of input and mechanism~\cite{sgouristsa:15:cure}]\label{post:ind_input_mechanism}
If $X \rightarrow Y$, the marginal distribution of the cause $P(X)$, and the conditional distribution of the effect given the cause, $P(Y \mid X)$ are \emph{independent} --- $P(X)$ contains no information about $P(Y \mid X)$ --- and vice versa since they correspond to independent mechanisms of nature.
\end{postulate}

This postulate provides the foundation for many successful causal inference frameworks designed for a pair of variables~\cite{janzing:10:justifyanm,janzing:12:igci,sgouristsa:15:cure,schoelkopf:12:clearn}. We can think of conditional $P(Y|X)$ as the \emph{mechanism} that transforms $x$-values into $y$-values, i.e.\ generates effect $Y$ for cause $X$. The postulate is justified if we are dealing with a mechanism of nature that does not care what input we provide to it ($P(X)$ in this case).  This independence will not hold in the other direction as $P(Y)$ and $P(X \mid Y)$ may contain information about each other as both inherit properties from $P(Y \mid X)$ and $P(X)$. This creates an asymmetry between cause and effect.

It is insightful to consider the following example where amount of radiation per $\mathit{cm}^2$ solar cell (cause) causes power generation in the cell (effect). We can just affect $P(\cause)$ only by actions such as moving the solar cell to a shady place, and varying the angle to the sun to affect $P(\cause)$. Likewise we can change only $P(\effect \mid \cause)$ by actions such as using more efficient cells. However it is hard to find actions that change $P(\effect)$ without affecting $P(\cause \mid \effect)$ or vice versa.

The notion of \emph{independence}, however, is abstract. Accordingly, different formalisations have been proposed. Janzing et al.~\cite{janzing:12:igci} define independence in terms of information geometry. Liu \& Chan~\cite{liu:16:dc} formulate independence in terms of the distance correlation between marginal and conditional empirical distribution. Janzing \& Sch{\"o}lkopf~\cite{janzing:10:algomarkov} formalise independence using algorithmic information theory, and postulate \emph{algorithmic} independence of $P(X)$ and $P(Y \mid X)$. Since algorithmic formulation captures all types of dependencies, and has a sound theoretical foundation, it is, arguably, a better mathematical formalisation of Postulate~\ref{post:ind_input_mechanism}. Using algorithmic information theory, we arrive at the following postulate.

\begin{postulate}[algorithmic independence of Markov kernels~\cite{janzing:10:algomarkov}]\label{post:ind_algo}
If $\xtoy$, the marginal distribution of the cause $P(X)$ and the conditional distribution of the cause given the effect $P(Y \mid X)$ are algorithmically independent, i.e.\ $I(P(X):P(Y \mid X)) \peq 0$.
\end{postulate}

Postulate~\ref{post:ind_algo} is equivalent to saying that if $\xtoy$, factorizing the joint distribution over $X$ and $Y$ into $P(X)$ and $P(Y \mid X)$, will lead to simpler --- in terms of Kolmogorov complexity --- models than factorizing it into $P(Y)$ and $P(X \mid Y)$~\cite{janzing:10:algomarkov}. The following theorem is hence a consequence of the algorithmic independence of input and mechanism.
\begin{theorem}[Th.~1 in~\citet{mooij:10:latent}]\label{thm:joint_desc}
If $X$ is a cause of $Y$, 
\[
K(P(X)) + K(P(Y \mid X)) \leq K(P(Y)) + K(P(X \mid Y)) \; .
\]
holds up to an additive constant.
\end{theorem}
In other words, we can perform causal inference simply by identifying that direction between $X$ and $Y$ for which the 
factorization of the joint distribution has the lowest Kolmogorov complexity. Although this inference rule has sound theoretical foundations, the problem remains that Kolmogorov complexity is not computable because of the widely known \emph{halting problem}. In practice, we therefore need other, computable, notions of independence or information. We can, for instance, approximate Kolmogorov complexity from above through lossless compression~\cite{vitanyi:93:book}. More generally, the Minimum Description Length (MDL) principle~\cite{rissanen:78:mdl,grunwald:07:book} provides a statistically sound and computable means for approximating Kolmogorov complexity~\cite{vereshchagin:03:kolmo,grunwald:07:book}.

\section{Causal Inference by Compression}\label{sec:practical}
In this section, we discuss how stochastic complexity can be used for practical causal inference. We gradually move towards that goal starting with MDL, and covering the basics along the way.

\subsection{Minimum Description Length Principle}
The Minimum Description Length (MDL)~\cite{rissanen:78:mdl} principle is a practical version of Kolmogorov complexity. 
Instead of all possible programs, it considers only programs for which we know they generate $x$ and halt. That is, lossless compressors.

In MDL theory, programs are often referred to as \emph{models}. The MDL principle has its root in the two-part decomposition of the Kolmogorov complexity~\cite{vitanyi:93:book}. It can be roughly described as follows~\cite{grunwald:07:book}. Given a set of models $\mclass$ and data $D$, the best model $M \in \mclass$ is the one that minimises $L(D, M) = L(M)+L(D \mid M)$, where $L(M)$ is the length, in bits, of the description of the model, and $L(D \mid M)$ is the length, in bits, of the description of the data when encoded with the model $M$. Intuitively $L(M)$ represents the compressible part of the data, and $L(D \mid M)$ represents the noise in the data.

This is called two-part MDL, or \emph{crude} MDL. To use crude MDL in practice, we have to define our model class \mclass, and the description methods for $L(M)$ as well as $L(D \mid M)$. If the models \mclass under consideration define probability distributions, we can use optimal prefix code given by Shannon entropy, $L(D \mid M) = -\log P(D \mid M)$, where $P(D \mid M)$ is the probability mass or density function of $D$ according to $M$. The definition of $L(M)$, however, is tricky --- $L(M)$ can vary from one encoding to the other, introducing \emph{arbitrariness} in the process.

The \emph{refined} version of MDL overcomes this arbitrariness by encoding $M$ and $D$ together. Unlike crude MDL, refined MDL encodes $D$ with the (entire) model class \mclass, resulting in single \emph{one-part} code $\bar{L}(D \mid \mclass)$~\cite{grunwald:07:book}. The one-part code length $\bar{L}(D \mid \mclass)$ is also called the \emph{stochastic complexity} of $D$ with respect to $M$.

The code is designed in such a way that if there exists a model $M^* \in \mclass$ for which $L(D \mid M^*)$ is minimal then $\bar{L}(D \mid \mclass)$ will also be minimal. Codes with such property are also called \emph{universal codes}. There exist various types of universal codes. Although the coding schemes are different across those codes, the resulting code lengths $\bar{L}(D \mid \mclass)$ are almost the same~\cite{grunwald:07:book}. In this work, we consider the NML universal code in particular.


Next we explain stochastic complexity in detail using the NML universal code.

\subsection{Stochastic Complexity}
Let $X^n = (x_1, x_2, \dots, x_n)$ be an i.i.d.\ sample of $n$ observed outcomes, where each outcome $x_i$ is an element of a space of observations \spacex. Let $\paramsspace \in \mathrm{R}^d$, where $d \in \mathbb{Z}^+$, be the parameter space. A model class \mclass is a family of probability distributions consisting of all the different distributions $P(.\mid \params)$ that can be produced by varying the parameters \params. Formally, a model class $\mclass$ is defined as
\[
\mclass = \{ P(\cdot \mid \params) : \params \in \paramsspace \} \; .
\]
To encode the data $X^n$ optimally with respect to the model class \mclass, we can use the code corresponding to the distribution $P(\cdot \mid \hat{\params}(X^n, \mclass))$ induced by the maximum likelihood estimate $\hat{\params}(X^n, \mclass)$ of the data $X^n$ for a given model class \mclass, since this distribution assigns shorter code length, i.e.\ higher likelihood, to the data than any of the other distributions in the model class. The Normalized Maximum Likelihood (NML) distribution is then defined as
\[
\pnml{X^n \mid \mclass} = \frac{P(X^n \mid \hat{\params}(X^n, \mclass))}{R(\mclass, n)} \; ,
\]
where the normalizing term $R(\mclass, n)$ is the sum over maximum likelihoods of all possible datasets of size $n$ under the model class $\mclass$. For discrete data, $R(\mclass, n)$ is defined as
\[
R(\mclass, n) = \sum\limits_{Y^n \in \spacex^n} P(Y^n \mid \hat{\params}(Y^n, \mclass)) \; , \label{eq:normalizer}
\]
where $\spacex^n$ is the $n$-fold Cartesian product $\spacex \times \cdots \times \spacex$ indicating set of all possible datasets of size $n$ with domain \spacex. When the data $X^n$ is defined over a continuous sample space, the summation symbol in Equation~\ref{eq:normalizer} is replaced by an integral.

The NML distribution has a number of important theoretical properties. First, it gives a unique solution to the minimax problem posed by Shtarkov~\cite{shtarkov:87:universal},
\[
\min_{\hat{P}} \max_{X^n} \log \frac{P(X^n \mid \hat{\params}(X^n, \mclass))}{\hat{P}(X^n \mid \mclass)} \; .
\]
That is, for \emph{any} data $X^n$, $\pnml{X^n \mid \mclass}$ assigns a probability, which differs from the highest achievable probability within the model class --- the maximum likelihood $P(X^n \mid \hat{\params}(X^n, \mclass))$ --- by a constant factor $R(\mclass, n)$. In other words, the NML distribution is the \emph{mini-max optimal universal model} with respect to the model class~\cite{myung:06:nmltut}. The NML distribution represents the behaviour of all the distributions in the model class \mclass.

Second, it also provides solution to another mini-max problem formulated by Rissanen~\cite{rissanen:01:optregret}, which is given by
\[
\min_{\hat{P}} \max_{Q} E_{Q} \log \frac{P(X^n \mid \hat{\params}(X^n, \mclass))}{\hat{P}(X^n \mid \mclass)} \; ,
\]
where $Q$ is the worst-case data generating distribution, and $E_{Q}$ is the expectation over $X^n$. That is, even if the true data generating distribution does not reside in the model class \mclass under consideration,  $\pnml{X^n \mid \mclass}$ still gives the optimal encoding for the data $X^n$ relative to \mclass.

These properties are very important and relevant when modelling real-world problems. In most cases, we do not know the true data generating distribution. In such cases, ideally we would want to encode our data as best as possible --- close to the optimal under the true distribution. The NML distribution provides a theoretically sound means for that.


The \emph{stochastic complexity} of data $X^n$ relative to a model class \mclass using the NML distribution is defined as
\begin{align}
\scomp{X^n \mid \mclass} &= -\log \pnml{X^n \mid \mclass} \nonumber \\
						 &= -\log P(X^n \mid \hat{\params}(X^n, \mclass)) + \log R(\mclass, n) \; .\label{eq:sc}
\end{align}
The term $\log R(\mclass, n)$ is the \emph{parametric} complexity of the model class \mclass. It indicates how well \mclass can fit random data.

The stochastic complexity of data under a model class \mclass gives the shortest description of the data relative to \mclass. Hence the richer the \mclass, the closer we are to Kolmogorov complexity. Intuitively, it is also the amount of information, in bits, in the data relative to the model class. Moreover, it is evident from the formulation that the stochastic complexity of data, relative to a model class, depends only on the data and the model class, but not on the particular way the models are specified.


\subsection{Causal Inference by Stochastic Complexity}

Unless stated otherwise, we write $X$ for $X^n$, and $Y$ for $Y^n$. The stochastic complexity of data $X$ relative to model class \mclass corresponds to the complexity of the NML distribution of the data relative to \mclass. This means we can use the stochastic complexity of $X$ as an approximation of the Kolmogorov complexity of $P(X)$. As such, it provides a general, yet computable, theoretically sound foundation for causal inference based on algorithmic information theory.

For ease of notation, wherever clear from context we write $\scomp{X}$ for $\scomp{X \mid \mclass}$. To infer the causal direction, we look over total stochastic complexity in two directions --- $X$ to $Y$ and vice versa. The total stochastic complexity from $X$ to $Y$, approximating $K(P(X)) + K(P(Y \mid X))$ is given by
\[
\scompxtoy = \scomp{X} + \scomp{Y \mid X}\; ,
\]
and that from $Y$ to $X$ is given by
\[
\scompytox = \scomp{Y} + \scomp{X \mid Y}\; .
\]
Following Theorem~\ref{thm:joint_desc}, using the above indicators we arrive at the following causal inference rules.
\begin{itemize}
	\item If $\scompxtoy < \scompytox$, we infer \xtoy.
	\item If $\scompxtoy > \scompytox$, we infer \ytox.
	\item If $\scompxtoy = \scompytox$, we are undecided.
\end{itemize}

That is, if describing $X$ and then describing $Y$ given $X$ is easier --- in terms of stochastic complexity --- than vice versa, we infer $X$ is likely the cause of $Y$. If it is the other way around, we infer $Y$ is likely the cause of $X$. If both ways of describing are the same, we remain undecided. We refer to this framework as \ourmethod, which stands for causal inference by stochastic complexity.

Causal inference using stochastic complexity has a number of powerful properties. First, unlike Kolmogorov complexity, stochastic complexity is computable. Second, the inference rule is generic in the sense that we are not restricted to one data type or distribution---we are only constrained by the model class $\mclass$ under consideration, yet by the mini-max property of NML we know that even if the data generating distribution is adversarial, we still identify the best encoding relative to $\mclass$. 

Next we discuss how can we instantiate \ourmethod for discrete data.

\subsection{Multinomial Stochastic Complexity}
We consider discrete random variable $X$ with $m$ values. Furthermore we assume that our data $X^n = (x_1, \dots, x_n)$ is multinomially distributed. The space of observations \spacex is then $\{1, 2, \dots, m\}$. The multinomial model class $\mnmclass$ is defined as
\[
\mnmclass = \{P(X \mid \params) : \params \in \paramsspace_m \} \; ,
\]
where \paramsspace is the simplex-shaped parameter space given by 
\[
\paramsspace_m = \{\params = (\theta_1, \dots ,\theta_m) : \theta_j \geq 0, \theta_1 + \cdots + \theta_m = 1 \} \; ,
\]
with $\theta_j = P(X=j \mid \params), j=1,\dots,m$.
The maximum likelihood parameters for a multinomial distribution are given by $\hat{\params}(X^n, \mnmclass) = (h_1/n, \dots, h_m/n)$, where $h_j$ is the number of times an outcome $j$ is seen in $X^n$. Then the distribution induced by the maximum likelihood parameters for $X^n$ under the model class \mnmclass is given by
\begin{align*}
P(X^n \mid \hat{\params}(X^n, \mnmclass)) &= \prod\limits_{i=1}^{n} P(x_i \mid \hat{\params}(X^n, \mnmclass))\\
&= \prod_{j=1}^{m} \left ( \frac{h_j}{n} \right ) ^{h_j} \; .
\end{align*}
The normalizing term $R(\mnmclass, n)$ is given by 
\begin{align}\label{eq:norm_mnsc}
R(\mnmclass, n) &= \sum\limits_{Y^n \in \spacex^n} P(Y^n \mid \hat{\params}(Y^n, \mnmclass)) \nonumber \\
&= \sum\limits_{h_1+\cdots+h_m=n} \frac{n!}{h_1!\cdots h_m!} \prod\limits_{j=1}^{m} \left ( \frac{h_j}{n} \right )^{h_j} \; .
\end{align}

Then the NML distribution for $X^n$ under the model class \mnmclass is given by
\begin{align*}
\pnml{X^n \mid \mnmclass} = \frac{\prod_{j=1}^{m} (h_j/n)^{h_j}}{R(\mnmclass, n)} \; .
\end{align*}
Then the stochastic complexity of $X^n$ for the model class \mnmclass is given by 
\begin{align}\label{eq:mnsc}
\scomp{X^n \mid \mnmclass} &= -\log \prod_{j=1}^{m} (h_j/n)^{h_j} + \log R(\mnmclass, n) \nonumber \\
& = \sum_{j=1}^{m} h_j ( \log n - \log h_j) + \log R(\mnmclass, n) \nonumber \\
&= n \log n - \sum_{j=1}^{m} h_j \log h_j + \log R(\mnmclass, n) \; .
\end{align}

\textbf{Computational Complexity ---}
We can compute the counts $h_j$ in $\bigo{n}$ by going through the data once. However, computing the normalizing sum (Equation~\ref{eq:norm_mnsc}), and hence the parametric complexity, is exponential in the number of values $m$. As a result, the computational complexity of the multinomial stochastic complexity (Equation~\ref{eq:mnsc}) is dominated by by computation time of the normalizing sum.

However, we can approximate the normalising sum up to a finite floating-point precision in \emph{sub-linear} time with respect to the data size $n$ given precomputed counts $h_i$~\cite{mononen:08:sublinearsc}. More precisely, the computational complexity of the sub-linear algorithm is $\bigo{\sqrt{dn}+m}$, where $d$ is the floating-point precision in digits. In the experiments we use $d=10$. Altogether we can compute the multinomial stochastic complexity in $\bigo{n}$.

\subsection{Computing Conditional Complexity}
So far we only discussed how to compute the stochastic complexity of data under a model class. For our purpose, we also need to compute the conditional stochastic complexity $\scomp{Y \mid X}$ and vice versa. Let $\scomp{Y \mid X=x}$ be the stochastic complexity of $Y$ conditioned on $X=x$. Then the conditional stochastic complexity $\scomp{Y \mid X}$ is the sum of $\scomp{Y \mid X=x}$ over all possible values of $X$.


Let \spacex be the domain of $X$. Then the stochastic complexity of $Y$ given $X$ is defined as
\begin{align*}
\scomp{Y \mid X} = \sum_{x \in \spacex} \scomp{Y \mid X = x} \; .
\end{align*}

\textbf{Computational Complexity ---} We can compute $\scomp{Y \mid X=x}$ in $\bigo{n}$. To compute the conditional stochastic complexity $\scomp{Y \mid X}$, we have to compute $\scomp{Y \mid X=x}$ over all $x \in \spacex$. Hence the computational complexity of conditional stochastic complexity is $\bigo{n|\spacex|}$. Likewise, for $\scomp{X \mid Y}$, we have $\bigo{n|\spacey|}$. 
Altogether the computational complexity of \ourmethod is $\bigo{n \max(|\spacex|, |\spacey|)}$.

\section{Related Work}\label{sec:rel}
Inferring causal direction from observational data is a challenging task due to the lack of controlled randomised experiments. However, it has also attracted quite a lot of attention over the years~\cite{pearl:00:book,spirtes:00:book,shimizu:06:anm,janzing:10:algomarkov}. Yet, most of the causal inference frameworks are built for continuous real-valued data.

Constraint-based approaches like conditional independence test~\cite{spirtes:00:book,pearl:00:book} are one of the widely used causal inference frameworks. However, they require at least three observed random variables. Therefore they cannot distinguish between \xtoy and \ytox as the factorization of the joint distribution $P(X, Y)$ is the same in both direction, i.e.\ $P(X) P(Y\mid X) = P(Y) P(X \mid Y)$.

In recent years, several methods have been proposed that exploit the sophisticated properties of the joint distribution. The linear trace method~\cite{janzing:10:ltr,zscheischler:11:letr} infers linear causal relations of the form $Y = AX$, where $A$ is the structure matrix that maps the cause to the effect, using the linear trace condition. The kernelized trace method~\cite{chen:13:ktr} can infer non-linear causal relations, but requires the causal relation to be deterministic, functional, and invertible. In contrast, we do not make any assumptions on the causal relation between the variables. 

One of the key frameworks for causal inference are the Additive Noise Models (ANMs)~\cite{shimizu:06:anm}. ANMs assume that the effect is a function of the cause and the additive noise that is independent of the cause. Causal inference is then done by finding the direction that admits such a model. Over the years, many frameworks for causal inference from real-valued data have been proposed using ANMs~\cite{shimizu:06:anm,hoyer:09:nonlinear,zhang:09:ipcm,peters:14:continuousanm}.

Algorithmic information theory provides a sound general theoretical foundation for causal inference~\cite{janzing:10:algomarkov}. The key idea is that if $X$ causes $Y$, the shortest description of the joint distribution $P(X, Y)$ is given by the separate descriptions of the distributions $P(X)$ and $P(Y \mid X)$~\cite{janzing:10:algomarkov}. It has also been used in justifying the additive noise model based causal discovery~\cite{janzing:10:justifyanm}. 

However, as Kolmogorov complexity is not computable, practical instantiations require computable notions of independence. For instance, the information-geometric approach~\cite{janzing:12:igci} defines independence via orthogonality in information space. \textsc{Cure}~\cite{sgouristsa:15:cure} defines independence in terms of the accuracy of the estimations of $P(Y \mid X)$ and $P(X \mid Y)$. Using algorithmic information theory, Vreeken~\cite{vreeken:15:ergo} proposes a causal framework based on relative conditional complexity and instantiates it with cumulative entropy to infer the causal direction in continuous real-valued data. Budhathoki \& Vreeken~\cite{budhathoki:16:origo} propose a decision tree based approach for causal inference on univariate and multivariate binary data.

All above methods consider either continuous real-valued or binary data. Causal inference from discrete data has received much less attention. Peters et al.~\cite{peters:10:discreteanm} (\dr) extend additive noise models to discrete data. However regression is not ideal for modelling categorical variables, and as it relies on the dependence measure, the choice of which affects the outcome. Liu \& Chan~\cite{liu:16:dc} (\dc) define independence in terms of the distance correlation between empirical distributions $P(X)$ and $P(Y \mid X)$ to infer the causal direction from categorical data. As such, it does not look over all possible space of the observed samples and hence overfits.

In contrast, we look over all possible space of the observed samples. Moreover, we provide a general, yet computable, theory for causal inference that is applicable to any type of data. In particular, we directly approximate Kolmogorov complexity using a score that is mini-max optimal with regard to the model class under consideration. The computational complexity of our instantiation, \ourmethod, is linear in sample size, regardless of the domain of the variables. In the experiments, we consider both \dc and \dr for comparison.

\section{Experiments}\label{sec:exps}
We implemented \ourmethod in Python and provide the source code for research purposes, along with the used datasets, and synthetic dataset generator.\!\footnote{\oururl} All experiments were executed single-threaded on Intel Xeon E5-2643 v3 machine with $256$GB memory running Linux. We consider synthetic, benchmark, and real-world data. In particular, we note that \ourmethod is parameter-free. We compare \ourmethod against Discrete Regression (\dr)~\cite{peters:10:discreteanm}, and \dc~\cite{liu:16:dc}. In particular, we use significance level of $\alpha=0.05$ for the independence test in \dr, and threshold of $\epsilon = 0.0$ for \dc.

\subsection{Synthetic Data}\label{subsec:synthetic_data}
To evaluate \ourmethod on the data with known ground truth, we consider synthetic data. Generating non-trivial synthetic data with identifiable causal direction is surprisingly difficult, though.\!\footnote{Ideally we would generate data with known $K(P(X))+K(P(Y\mid X)) < K(P(Y))+K(P(X\mid Y))$, and evaluate our inference methods accordingly, yet as Kolmogorov complexity is not computable it is not apparent how to do this in general.} 
We generate synthetic cause-effect pairs with ground truth \xtoy using the additive noise model (ANM). That is, first we generate the cause $X$, and then generate the effect $Y$ using the model given by
\[
Y = f(X) + N, N \indep X \; ,
\]
where $f$ is a function, and $N$ is additive noise that is independent of $X$. Following \citet{peters:10:discreteanm}, we sample $X$ from the following distributions, using independently generated uniform noise.
\begin{itemize}
\item uniform from $\{1, \dots, L\}$,
\item binomial with parameters $(n, p)$,
\item geometric with parameter $p$,
\item hypergeometric with parameters $(M, K, N)$,
\item poisson with parameter $\lambda$,
\item negative binomial with parameters $(n, p)$, and
\item multinomial with parameters $\params$.
\end{itemize}
We note that even though we generate data following ANM from $X$ to $Y$, the joint distribution $P(X, Y)$ might admit an additive noise model in the reverse direction. Therefore in some cases where we say that \xtoy is the true direction, \ytox might also be equally plausible, and hence full accuracy might not be achievable in some cases. However, this happens in only few trivial instances~\cite{peters:10:discreteanm}.

We choose parameters of the distributions randomly for each model class. We choose $L$ uniformly between $1$ and $10$, $M, K$ uniformly between $1$ and $40$, $N$ uniformly between $1$ and $\min(41, M + K)$, $p$ uniformly between $0.1$ and $0.9$, $\lambda$ uniformly between $1$ and $10$, $\params$ randomly s.t. $\sum_{\theta \in \params} \theta = 1.0$, function $f(x)$ uniformly between $-7$ to $+7$, and noise $N$ uniformly between $-t$ to $+t$, where $t$ is uniformly randomly chosen between $1$ and $7$.

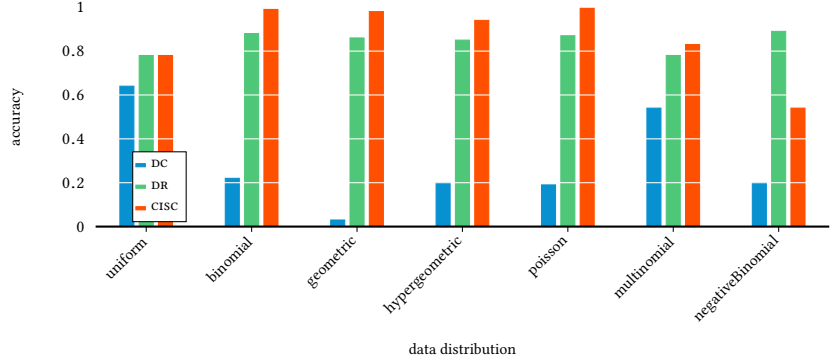
\begin{figure}[t]
	\centering
\ifgenplot
	\ifpdf
	\tikzsetnextfilename{accuracy_synth}
	\fi
	\begin{tikzpicture}
		\begin{axis}[eda ybar, 
		width=0.75\columnwidth, height=4.5cm,
		ymin=0.0, ymax=1.0, enlarge x limits=0.08, 
		bar width=0.59em, 
		xlabel=data distribution, 	
		ylabel=accuracy, 
		x label style={yshift=-20pt},
		y label style={yshift=-25pt},
		xtick=data,  
		symbolic x coords={uniform,binomial,geometric,hypergeometric,poisson,multinomial,negativeBinomial}, 
		x tick label style={rotate=45, anchor=east, align=right, yshift=-2pt}, 
		legend style={anchor=south, at={(0.087,0.02)}}, legend image post style={scale=0.6}, 
		]
		\pgfplotsinvokeforeach{1,...,3}{
			\addplot table[x index=0, y index=#1, header=false] {../expres/accuracy_synthetic.dat};
		}
		\legend{\dc, \dr,\ourmethod};
		\end{axis}
	\end{tikzpicture}
\else
	\includegraphics[]{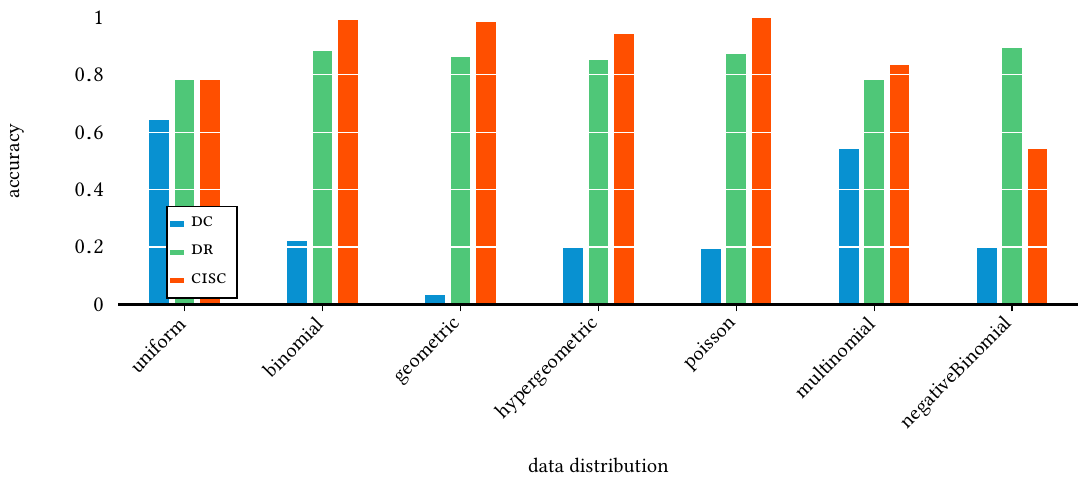}
\fi
	\caption{Accuracy on synthetic cause-effect pairs sampled from different distributions.}
	\label{fig:acc_synth_data}
\end{figure}

\textbf{Accuracy ---} From each model class, we sample $1000$ different models, and hence $1000$ different cause-effect pairs. For each model, we sample $1000$ points, i.e.\ $n=1000$. In Figure~\ref{fig:acc_synth_data}, we compare the \emph{accuracy} (percentage of correct decisions) of \ourmethod against \dc and \dr for various model classes. We see that \ourmethod either outperforms or is as good as the other methods in all but one case. This certainly proves the generality of \ourmethod. 

Although we compute the stochastic complexity under multinomial model class, we are still able to perform as good with other model classes. This is due to the optimality property of the NML distribution -- even though the true data generating distribution is not inside the model class \mclass under consideration, the NML distribution still gives the optimal encoding relative to \mclass. And as we see, it works well in most cases.

\textbf{Decision Rate ---}
Next we investigate the accuracy of \ourmethod against the fraction of decisions \ourmethod is forced to make. To this end, for each model class, we sample $1000$ different cause-effect pairs. For each cause-effect pair, we sample $1000$ points. We sort the pairs by their absolute score difference in two directions (\xtoy vs. \ytox), i.e.\ $|\scompxtoy - \scompytox|$ in descending order. Then we compute the accuracy over top-$k\%$ pairs. The decision rate is the fraction of \emph{top} cause-effect pairs that we consider. Alternatively, it is also the fraction of cause-effect pairs whose $|\scompxtoy - \scompytox|$ is greater than some threshold $\delta$. For undecided pairs, we flip the coin. For other methods, we follow the similar procedure with their respective absolute score difference. 

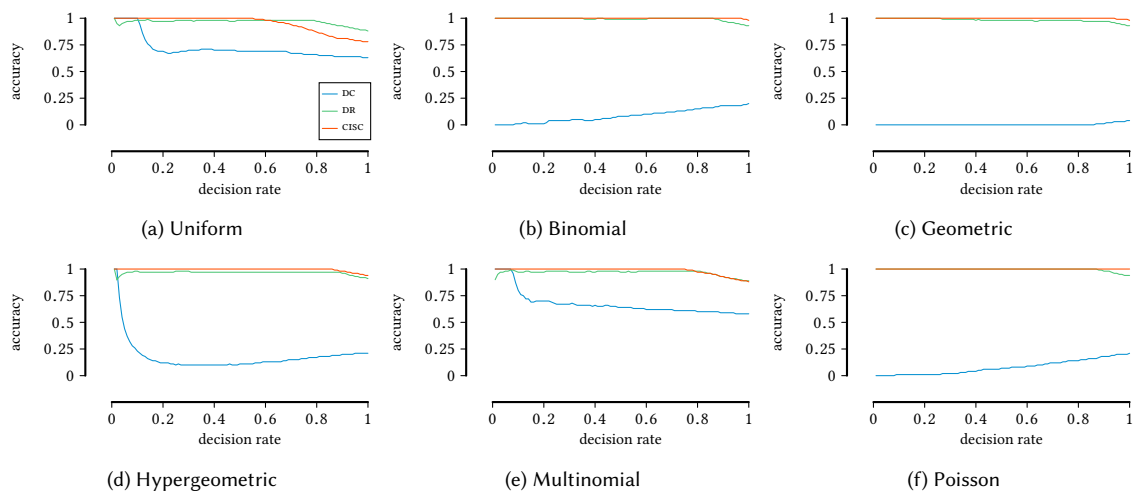
\begin{figure*}[tb]
	\begin{minipage}[t]{0.33\linewidth}
		\centering
		\ifgenplot
			\ifpdf
			\tikzsetnextfilename{dec_rate_synth_uniform}
			\fi
			\begin{tikzpicture}
			\begin{axis}[eda line, 
			height=3cm, width=\textwidth,
			xlabel=decision rate,ylabel=accuracy, ymin=0.0, ymax=1.0, xmin=0.0, xmax=1.0, legend style={nodes={scale=0.8, transform shape}, at={(0.91,0.4)}, anchor=north}, ytick={0.0,0.25, ...,1.0}]
			\pgfplotsinvokeforeach{1,...,3}{
				\addplot table[x index=0, y index=#1, header=false] {../expres/dec_rate_synth_uniform.dat};
			}
			\legend{\dc, \dr, \ourmethod}
			\end{axis}
			\end{tikzpicture}
		\else
			\includegraphics[]{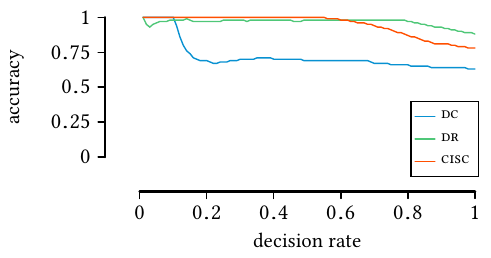}
		\fi
		\vspace{-0.4cm}
		\subcaption{Uniform}\label{fig:dec_rate_uniform}
	\end{minipage}
	\begin{minipage}[t]{0.33\linewidth}
		\centering
		\ifgenplot
			\ifpdf
			\tikzsetnextfilename{dec_rate_synth_binom}
			\fi
			\begin{tikzpicture}
			\begin{axis}[eda line, xlabel=decision rate,ylabel=accuracy, ymin=0.0, ymax=1.0, height=3cm, width=\textwidth, xmin=0.0, xmax=1.0, ytick={0.0,0.25, ...,1.0}]
			\pgfplotsinvokeforeach{1,...,3}{
				\addplot table[x index=0, y index=#1, header=false] {../expres/dec_rate_synth_binomial.dat};
			}
			\end{axis}
			\end{tikzpicture}
		\else
			\includegraphics{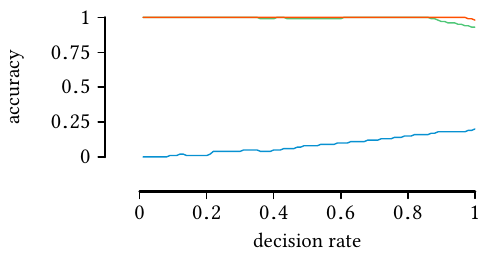}
		\fi
		\vspace{-0.4cm}
		\subcaption{Binomial}\label{fig:dec_rate_binomial}
	\end{minipage}
	\begin{minipage}[t]{0.33\linewidth}
		\centering
		\ifgenplot
			\ifpdf
			\tikzsetnextfilename{dec_rate_synth_geometric}
			\fi
			\begin{tikzpicture}
			\begin{axis}[eda line, xlabel=decision rate,ylabel=accuracy, ymin=0.0, ymax=1.0, height=3cm, width=\textwidth, xmin=0.0, xmax=1.0, legend style={nodes={scale=0.8, transform shape}, at={(0.93,0.3)}, anchor=north},ytick={0.0,0.25, ...,1.0}]
			\pgfplotsinvokeforeach{1,...,3}{
				\addplot table[x index=0, y index=#1, header=false] {../expres/dec_rate_synth_geometric.dat};
			}
			\end{axis}
			\end{tikzpicture}
		\else
			\includegraphics{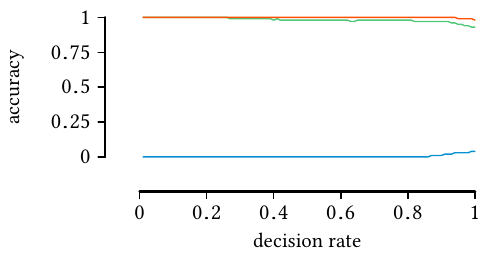}
		\fi
		\vspace{-0.4cm}
		\subcaption{Geometric}\label{fig:dec_rate_geometric}
	\end{minipage}
	\par\medskip
	\begin{minipage}[t]{0.33\linewidth}
		\centering
		\ifgenplot
			\ifpdf
			\tikzsetnextfilename{dec_rate_synth_hypergeo}
			\fi
			\begin{tikzpicture}
			\begin{axis}[eda line, xlabel=decision rate,ylabel=accuracy, ymin=0.0, ymax=1.0, height=3cm, width=\textwidth, xmin=0.0, xmax=1.0, ytick={0.0,0.25, ...,1.0}]
			\pgfplotsinvokeforeach{1,...,3}{
				\addplot table[x index=0, y index=#1, header=false] {../expres/dec_rate_synth_hypergeometric.dat};
			}
			\end{axis}
			\end{tikzpicture}
		\else
			\includegraphics{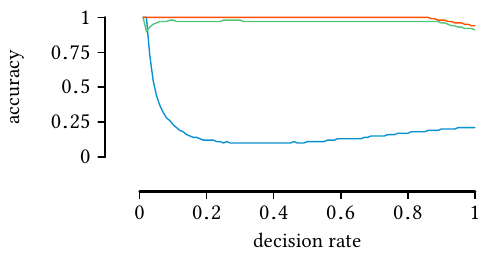}
		\fi
		\vspace{-0.4cm}
		\subcaption{Hypergeometric}\label{fig:dec_rate_hypergeometric}
	\end{minipage}
	\begin{minipage}[t]{0.33\linewidth}
		\centering
		\ifgenplot
			\ifpdf
			\tikzsetnextfilename{dec_rate_synth_multinom}
			\fi
			\begin{tikzpicture}
			\begin{axis}[eda line, xlabel=decision rate,ylabel=accuracy, ymin=0.0, ymax=1.0, height=3cm, width=\textwidth, xmin=0.0, xmax=1.0, ytick={0.0,0.25, ...,1.0}]
			\pgfplotsinvokeforeach{1,...,3}{
				\addplot table[x index=0, y index=#1, header=false] {../expres/dec_rate_synth_multinomial.dat};
			}
			\end{axis}
			\end{tikzpicture}
		\else
			\includegraphics{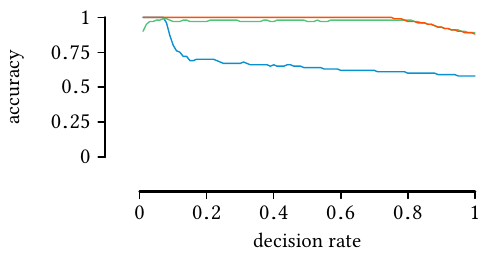}
		\fi
		\vspace{-0.4cm}
		\subcaption{Multinomial}\label{fig:dec_rate_multinomial}
	\end{minipage}
	\begin{minipage}[t]{0.33\linewidth}
		\centering
		\ifgenplot
			\ifpdf
			\tikzsetnextfilename{dec_rate_synth_poisson}
			\fi
			\begin{tikzpicture}
			\begin{axis}[eda line, xlabel=decision rate,ylabel=accuracy, ymin=0.0, ymax=1.0, height=3cm, width=\textwidth, xmin=0.0, xmax=1.0, ytick={0.0,0.25, ...,1.0}]
			\pgfplotsinvokeforeach{1,...,3}{
				\addplot table[x index=0, y index=#1, header=false] {../expres/dec_rate_synth_poisson.dat};
			}
			\end{axis}
			\end{tikzpicture}
		\else
			\includegraphics{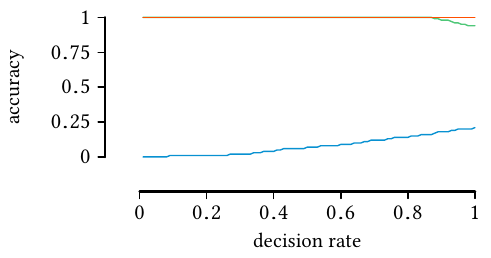}
		\fi
		\vspace{-0.4cm}
		\subcaption{Poisson}\label{fig:dec_rate_poisson}
	\end{minipage}
	\caption{Accuracy against decision rate on synthetic cause-effect pairs sampled from different distributions.} \label{fig:dec_rate_synthetic}
\end{figure*}

In Figure~\ref{fig:dec_rate_synthetic}, we show the decision rate versus accuracy for different model classes. We see that both \ourmethod and \dr are highly accurate up to a very high decision rate in all cases. Both \ourmethod and \dr are highly accurate on the cause-effect pairs where the absolute score difference is very high --- where the methods are most decisive. \dc, on the other hand, doesn't perform well in all cases. The only setting where $\dc$ has a relatively good performance is in the family of Uniform distributions. 

The results indicate that we can increase the threshold $\delta$, and hence the decision rate, for higher accuracy.

\textbf{Scalability ---}
Next we empirically investigate the scalability of \ourmethod. First, we examine runtime with regard to the sample size. To this end, we fix the domain size of the cause-effect pairs to $20$, i.e.\ $|\spacex| = |\spacey| = 20$. Then for a given sample size, we sample $X$ uniformly randomly between $1$ and $|\spacex|$. Likewise for $Y$.

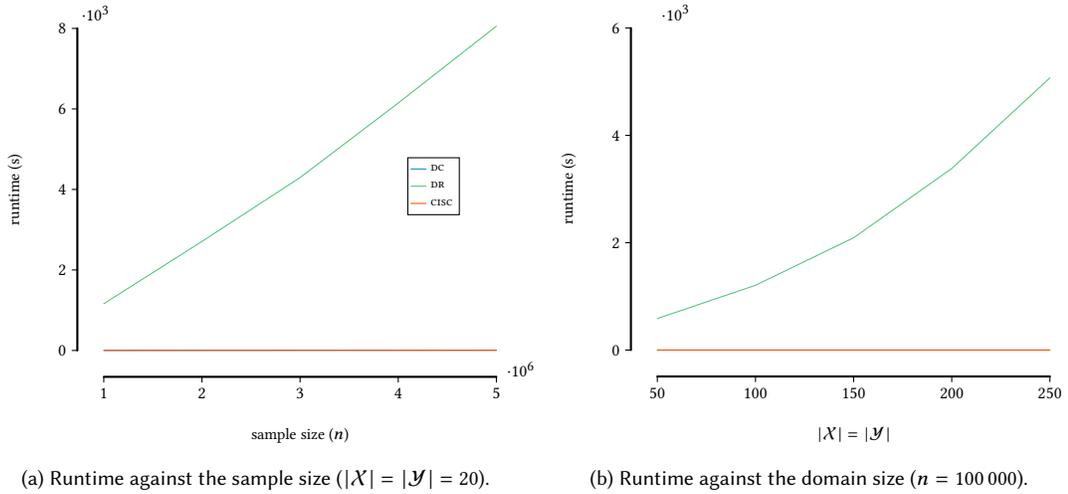
\begin{figure}[tb]
	\begin{minipage}[t]{0.45\columnwidth}
		\centering
		\ifgenplot
			\ifpdf
			\tikzsetnextfilename{size_vs_runtime}
			\fi
			\begin{tikzpicture}
			\begin{axis}[eda line, 
			ylabel={runtime (s)},
			xlabel={sample size ($n$)},
			width=\textwidth, 
			legend style={nodes={scale=0.8, transform shape}, at={(0.84,0.6)}, anchor=north}, 
			y label style={yshift=0pt}, 
			scaled x ticks=base 10:-6, 
			x tick scale label style={xshift=15pt, yshift=10pt},
			scaled y ticks=base 10:-3, 
			y tick scale label style={xshift=-10pt}, 
			ytick={0, 2000, ..., 8000}, 
			ymax=8000
			]
			\pgfplotsinvokeforeach{1,...,3}{
				\addplot table[x index=0, y index=#1, header=false] {../expres/size_vs_runtime.dat};
			}
			\legend{\dc, \dr, \ourmethod}
			\end{axis}
			\end{tikzpicture}
		\else
			\includegraphics{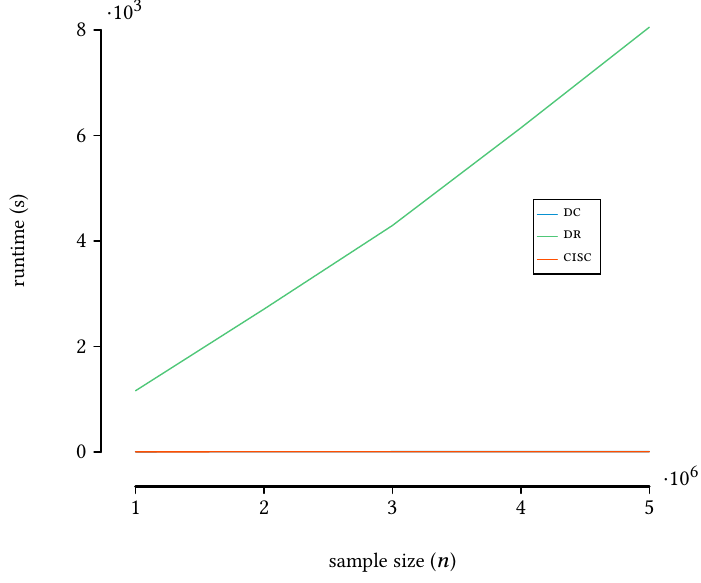}
		\fi
		\subcaption{Runtime against the sample size ($|\spacex|=|\spacey|=20$).}
		\label{fig:size_vs_runtime}
	\end{minipage}
	\hspace{0.4cm}
	\begin{minipage}[t]{0.45\columnwidth}
		\centering
		\ifgenplot
			\ifpdf
			\tikzsetnextfilename{domain_vs_runtime}
			\fi
			\begin{tikzpicture}
			\begin{axis}[eda line, 
			ylabel={runtime (s)},
			xlabel={$|\spacex|=|\spacey|$}, 
			width=\textwidth, 
			y label style={yshift=0pt}, 
			ytick={0, 2000, ..., 6000}, 
			ymax=6000,
			scaled y ticks=base 10:-3, 
			y tick scale label style={xshift=0pt}]
			\pgfplotsinvokeforeach{1,...,3}{
				\addplot table[x index=0, y index=#1, header=false] {../expres/domain_vs_runtime.dat};
			}		
			\end{axis}
			\end{tikzpicture}
		\else
			\includegraphics{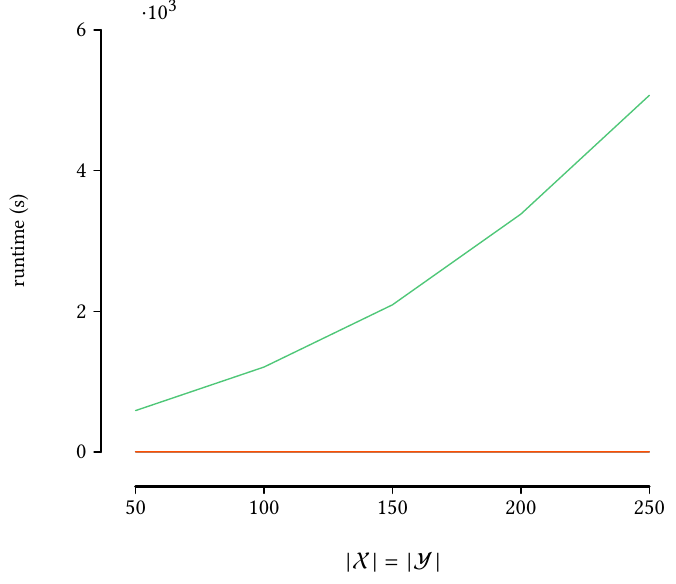}
		\fi
		\subcaption{Runtime against the domain size ($n = 100\,000$).}
		\label{fig:domain_vs_runtime.dat}
	\end{minipage}
	\caption{Runtime of the frameworks against (a) sample size, and (b) domain size.}
\end{figure}

In Figure~\ref{fig:size_vs_runtime}, we show the runtime of \ourmethod, \dc, and \dr for various sample sizes. We observe that both \ourmethod and \dc (overlapping line) finish within seconds. \dr, on the other hand, takes in the order of hours.

Next we fix the sample size to $n = 100\,000$ and vary the domain size $|\spacex|=|\spacey|$. We observe that both \ourmethod and \dc again finish within seconds over the whole range. As \dr iteratively searches over the entire domain, it shows a non-linear runtime behaviour with respect to the domain size.

Overall, these results indicate that \dr is fairly accurate, but relatively slow. 
\dc, on the other hand, is fast, yet inaccurate. 
\ourmethod is both highly accurate, and fast.

\subsection{Benchmark Data}
Next we evaluate \ourmethod on benchmark cause-effect pairs with known ground truth~\cite{mooij:16:pairs}. In particular, we take $95$ univariate cause-effect pairs. So far there does not exist a discretization strategy that provably preserves the causal relationship between variables. Since each cause-effect pair is from a different domain, using one discretization strategy over all the pairs is also unfair. Moreover, we do not know the underlying domain of the data. As a result, we treat the data as discrete for all the pairs.

\begin{figure}[tb]
	\centering
	\ifgenplot
		\ifpdf
		\tikzsetnextfilename{dec_rate_benchmark}
		\fi
		\begin{tikzpicture}
		\begin{axis}[eda line, xlabel=decision rate,ylabel=accuracy, ymin=0.0, ymax=1.0, 
		width=0.75\columnwidth, height=4.1cm, 
		y label style={yshift=5pt}, 		
		xmin=0.0, xmax=1.0, legend style={nodes={scale=0.8, transform shape}, at={(0.94,0.22)}, anchor=north}]
		\pgfplotsinvokeforeach{1,...,3}{
			\addplot table[x index=0, y index=#1, header=false] {../expres/dec_rate_benchmark.dat};
		}
		
		\addplot[name path=f, color=verylightgray] table[x index = 0, y index = 7, header = false] {../expres/dec_rate_crack.dat};
		\addplot[name path=g, color=verylightgray] table[x index = 0, y index = 8, header = false] {../expres/dec_rate_crack.dat};
		\addplot [verylightgray ] fill between[of=f and g, soft clip={domain=0:1}];
		\legend{\dc, \dr, \ourmethod}
		\end{axis}
		\end{tikzpicture}
	\else
		\includegraphics{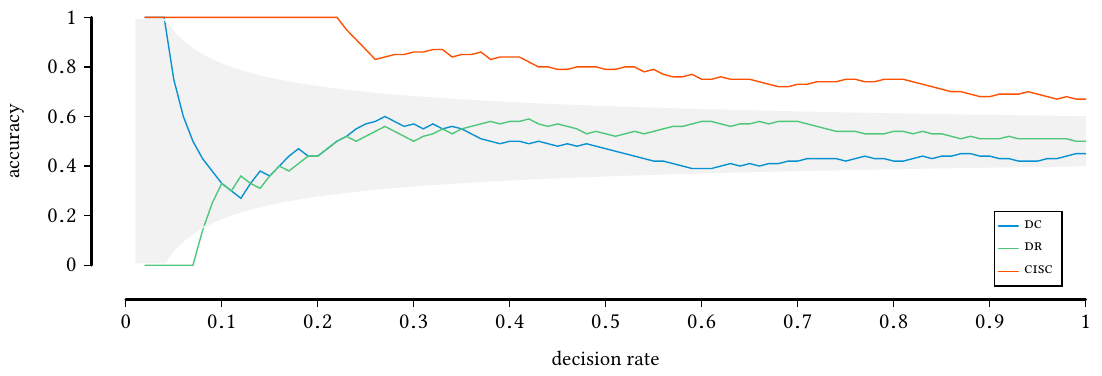}
	\fi
	\caption{Accuracy against decision rate for univariate T{\"u}bingen cause-effect pairs. Gray area indicates the 95\% confidence interval for a random coin flip.}
	\label{fig:dec_rate_benchmark}
\end{figure}
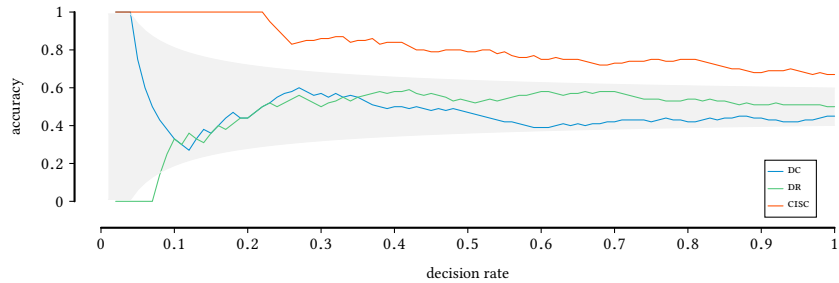

In Figure~\ref{fig:dec_rate_benchmark}, we compare the accuracy of \ourmethod against \dc and \dr at various decision rate together with the $95\%$ confidence interval for a random coin flip. If we look over all the pairs, we find that \ourmethod infers correct direction in roughly $67\%$ of all the pairs. When we consider only those pairs where \ourmethod is most decisive---with a very high value of $|\scomp{\xtoy} - \scomp{\ytox}|$, it is $100\%$ accurate on top $22\%$ of the pairs, $80\%$ accurate on top $45\%$ of the pairs, which is on-par with the top-performing causal inference frameworks for continuous real-valued data~\cite{sgouristsa:15:cure,janzing:12:igci}. On the other hand, the results from both \dc and \dr are insignificant at almost every decision rate.

\subsection{Qualitative Case Studies}
Next we evaluate \ourmethod on real-world data for exploratory purpose.

\textbf{Abalone ---} First we consider the \emph{Abalone} dataset, which is available from the UCI machine learning repository.\!\footnote{\label{fnote:uci}\url{http://archive.ics.uci.edu/ml/}} The dataset contains the physical measurements of $4\,177$ abalones, which are large, edible sea snails. 

Out of the nine measurements, we consider the \emph{sex} ($X$), \emph{length} ($Y_1$), \emph{diameter} ($Y_2$), and \emph{height} ($Y_3$). The length, diameter, and height of the abalone are all measured in millimetres, and have $70$, $57$ and $28$ different values, respectively whereas the sex of the abalone is nominal ($\text{male}=1$, $\text{female}=2$, or $\text{infant}=3$). Following Peters et al.~\cite{peters:10:discreteanm}, we regard the data as discrete, and consider $X \rightarrow Y_1$, $X \rightarrow Y_2$, and $X \rightarrow Y_3$ as the ground truth as sex causes the size of the abalone and not the other way around. \ourmethod infers correct direction in all three cases.

\textbf{Car Evaluation ---} The Car Evaluation dataset is available from the UCI machine learning repository. It has $1728$ rows, and is derived from a hierarchical decision model. It contains the evaluation of a car for buying purpose based on six characteristics of the car. 

We consider the estimated \emph{safety} ($X$) of the car against the \emph{evaluation} ($Y$) of the car. The safety feature of the car takes a nominal value ($\text{low}=1$, $\text{medium}=2$, or $\text{high}=3$), and the evaluation feature of the car also takes a nominal value ($\text{unacceptable}=1$, $\text{acceptable}=2$, $\text{good}=3$, or $\text{very good}=4$). We regard \xtoy as the ground truth as safety of the car causes the decision on buying the car, but not vice versa. \ourmethod identifies the correct direction.

\textbf{Adult ---} The \emph{Adult} dataset is taken from the UCI machine learning repository and consists of $48\,832$ records from the census database of the US in $1994$. 

Out of $14$ attributes, we consider only three -- \emph{education} ($X_1$), \emph{occupation} ($X_2$), and \emph{income} ($Y$). The domain of \emph{education} attribute consists of \emph{dropout, associates, bachelors, doctorate, hs-graduate, masters}, and \emph{prof-school}. For \emph{occupation}, we have \emph{admin, armed-force, blue-collar, white-collar, service, sales, professional}, and \emph{other-occupation} as possible values. Lastly, for \emph{income} attribute, we have two values: \emph{\textgreater 50K} and \emph{\textless=50}.

As intuitively education causes income, and not vice versa, 
we regard $X_1 \rightarrow Y$ as the ground truth. 
Similarly, as occupation causes income, 
we regard $X_2 \rightarrow Y$ as the ground truth. 
We run \ourmethod on both pairs $(X_1, Y)$ and $(X_2, Y)$. We observe that for both pairs \ourmethod infers the causal direction correctly.

Overall, these results illustrate that \ourmethod finds sensible causal directions from real-world data.

\section{Discussion}\label{sec:disc}
The experiments show that \ourmethod works well in practice. \ourmethod reliably identifies true causal direction regardless of the data distribution. It is remarkably fast. On benchmark data, it's performance is comparable to the state-of-the-art causal inference frameworks for continuous real-valued data. Moreover, the qualitative case studies show that the results are sensible.

In this work, we give a general framework for causal inference based on the solid foundations of information theory. To apply the framework in practice, we just have to compute the stochastic complexity relative to a model class. The richer the model class, the better the solution. Although computing the stochastic complexity involves looking over all possible datasets, theoretically it is still computable, and there do exist efficient algorithms for certain model classes. The proposed framework lays a clear computable foundation for algorithmic causal inference principle postulated by Janzing \& Sch{\"o}lkopf~\cite{janzing:10:algomarkov}.

Although the results show the strength of the proposed framework, and of \ourmethod in particular, we see many possibilities to further improve. We instantiated the framework using multinomial stochastic complexity on discrete data. We see that \ourmethod performs relatively well even in cases where the data is not sampled from the multinomial model class. This is due to the optimality property of the multinomial distribution --- even if the true data generating distribution is not inside the model class \mclass under consideration, the NML distribution still gives the optimal encoding for the data relative to \mclass. It would be an engaging future work to instantiate the framework for other types of data (e.g. continuous real-valued, mixed, etc.) and model classes (e.g. family of Gaussians, Dirichlets, etc.). The key aspect to study would be efficient algorithms for computing the stochastic complexity for such model classes.

We define conditional stochastic complexity $\scomp{Y \mid X}$ as the sum of the stochastic complexities of $Y$ conditioned on $X=x$ over all $x$. This way we look over local stochastic complexities of parts of $Y$ relative to each value of $x$. Perhaps we can compute the conditional stochastic complexity globally relative to $X$. It would also be interesting to explore factorized normalized maximum likelihood models~\cite{roos:08:structure} to instantiate the framework for multivariate data~\cite{budhathoki:16:origo}. 

To infer the causal relationship between variables $X$ and $Y$, we assume that there is no confounding variable $Z$. It would be interesting to use the framework to additionally discover the confounding variables. The rough idea is that factorizing the joint complexity $P(X, Y)$ in presence of the confounding variable $Z$ leads to the smallest stochastic complexity compared to factorizing into $P(X)$ and $P(Y \mid X)$ or $P(Y)$ and $P(X \mid Y)$.
 
Another avenue for future work would be to use the framework for causal discovery. The proposed framework infers causal relationship between given two variables $X$ and $Y$. It would be interesting to explore how the framework can be employed to discover (mine) the causal models \emph{directly} from the data. 

\section{Conclusion}\label{sec:concl}
We considered causal inference from observational data. We proposed a general, yet \emph{computable} framework for information-theoretic causal inference with optimality guarantees.
In particular, we proposed to perform causal inference by stochastic complexity.

To illustrate the strength of this, we proposed \ourmethod for pairs of univariate discrete variables, using stochastic complexity over the class of multinomial distributions. Extensive evaluation on synthetic, benchmark, and real-world data showed that \ourmethod is highly accurate, outperforming the state of the art by a margin, and scales extremely well with regard to both sample and domain sizes.

Future work includes considering richer model classes, as well as structure learning for the discovery of causal models from data.

\begin{acks}
Kailash Budhathoki is supported by the International Max Planck Research School for Computer Science. 
Both authors are supported by the Cluster of Excellence ``Multimodal Computing and Interaction'' within the Excellence Initiative of the German Federal Government.
\end{acks}

\bibliographystyle{ACM-Reference-Format}
\bibliography{abbrev,bib-jilles,bib-cisc}

\ifapx
\appendix
\include{appendix}
\fi

\balance

\end{document}